\documentclass[conference]{IEEEtran}

\usepackage{graphicx}
\usepackage{amssymb}
\usepackage{multirow}
\usepackage{amsfonts}
\usepackage{footnote}
\usepackage{lipsum}
\usepackage[english]{babel}
\usepackage[backend=bibtex8]{biblatex}
\usepackage[utf8]{inputenc}
\usepackage{csquotes}
\usepackage{booktabs}
\usepackage{amsthm}
\usepackage{balance}
\usepackage{microtype}
\usepackage{hyperref}
\hypersetup{
    pdftitle={Temporally Evolving Community Detection and Prediction in Content-Centric Networks},
}

\usepackage[T1]{fontenc}
\usepackage[cmex10]{amsmath}
\usepackage{amssymb}
\usepackage[cmintegrals]{newtxmath}
\usepackage{bm}

\usepackage[usenames,dvipsnames]{color}

\newtheorem{definition}{Definition}
\makesavenoteenv{table}

\begin{document}

\title{Using link and content over time for embedding generation in Dynamic Attributed Networks}

\IEEEoverridecommandlockouts%

\IEEEpubid{\makebox[\columnwidth]{\color{red}Pre-print. Published at ECML-PKDD 2018. DOI: \href{https://doi.org/10.1007/978-3-030-10928-8_1}{10.1007/978-3-030-10928-8\_1}} \hspace{\columnsep}\makebox[\columnwidth]{}}

\author{
    \IEEEauthorblockN{Ana Paula Appel\IEEEauthorrefmark{1}, Renato L. F. Cunha\IEEEauthorrefmark{1}, Charu C. Aggarwal\IEEEauthorrefmark{1}, Marcela Megumi Terakado\IEEEauthorrefmark{2}\thanks{\IEEEauthorrefmark{2}Work done while at IBM Research.}}
    \IEEEauthorblockA{\IEEEauthorrefmark{1}IBM Research\\\texttt{\{apappel,renatoc\}@br.ibm.com, charu@us.ibm.com}}
    \IEEEauthorblockA{\IEEEauthorrefmark{2}University of São Paulo\\\texttt{terakado@ime.usp.br}}
}
    
\maketitle

\begin{abstract}
    In this work, we consider the problem of creating an embedding combining link, content and temporal analysis for community detection and prediction in evolving
    networks. Such temporal and content-rich networks occur in
    many real-life settings, such as bibliographic networks and question
    answering forums. Most of the work in the literature (that uses both
    content and structure) deals with static snapshots of
    networks, and they do not reflect the dynamic changes occurring over
    multiple snapshots.  Incorporating dynamic changes in the
    communities into the analysis can also provide useful insights about
    the changes in the network such as the migration of authors across
    communities. In this work, we propose \emph{Chimera}\footnote{\texttt{https://github.com/renatolfc/chimera-stf}}, 
    a shared factorization model
    that can simultaneously account for graph links, content, and
    temporal analysis. This approach works by extracting the latent semantic
    structure of the network in multidimensional form, but in a way that
    takes into account the temporal continuity of these embeddings. Such
    an approach simplifies temporal analysis of the underlying
    network by using the embedding as a surrogate. A consequence of this
    simplification is that it is also possible to use this temporal
    sequence of embeddings to predict future communities.  We present
    experimental results illustrating the effectiveness of the approach.
\end{abstract}

\section{Introduction}

Structural representations of data are ubiquitous in different domains such as
biological networks, online social networks, information networks,
co-authorship networks, and so on. The problem of community detection or graph
clustering aims to identify densely connected groups of nodes in the
network~\cite{fortunato2010community}, one of the central tasks in network
analysis. Examples of useful applications include that of finding clusters in
protein-protein interaction networks~\cite{ravasz2002hierarchical} or groups of
people with similar interests in social networks~\cite{watts2002identity}.
Recently, it has become easier to collect content-centric networks in
a time-sensitive way, enabling the possibility of using tightly-integrated
analysis across different factors that affect network structure. Aggregate
topological and content information can enable more informative community
detection, in which cues from different sources are integrated into more
powerful models.

Another important aspect of complex networks is that such networks evolve,
meaning that nodes may move from one community to another, making some
communities grow, and others shrink. For example, authors that usually publish
in the data mining community could move to the machine learning community.
Furthermore, the temporal aspects of changes in community structure could
interact with the content in unusual ways.  For example, it is possible for an
author in a bibliographic network to change their topic of work, preceding
a corresponding change in community structure. The converse is also possible,
with a change in community structure affecting content-centric attributes.

Matrix factorization methods are traditional techniques that allow us to reduce
the dimensional space of network adjacency representations. Such methods have
broad applicability in various tasks such as clustering, dimensionality
reduction, latent semantic analysis, and recommender systems. The main point of
matrix factorization methods is that they embed matrices in a \emph{latent
space} where the clustering characteristics of the data are often amplified.
A useful variant of matrix factorization methods is
\emph{shared} matrix factorization, which factors two or more different
matrices simultaneously. Shared matrix factorization is not new, and is used in
various settings where different matrices define different parts of the data
(e.g., links and content). This method could be used to embed link and content
in a shared feature space, which is convenient because it
allows the use of traditional clustering techniques, such as $k$-means.
However, incorporating a temporal aspect to the shared factorization process
adds some challenges concerning the adjustment of the shared factorization as
data evolves.

 \begin{figure}
 \centering
 \includegraphics[width=0.4\textwidth]{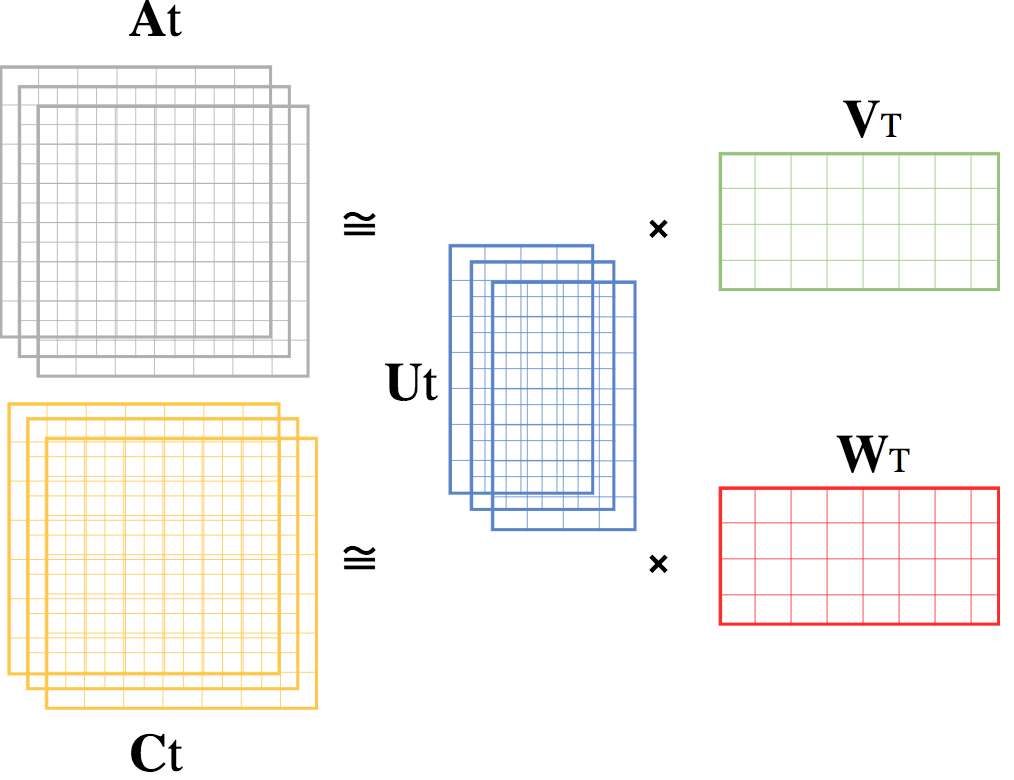}
 \caption{%
     Graphical representation how shared factorization on \emph{Chimera} works.
 }\label{fig:Shared}
 \end{figure}
 
\begin{figure}[htbp]
\centering
\includegraphics[width=0.45\textwidth]{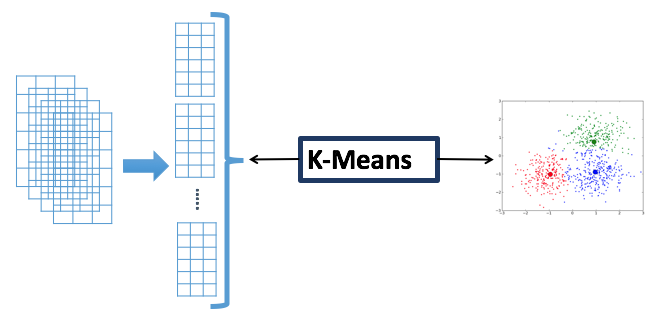}
\caption{%
    The different matrices corresponding to  $U$ are condensed in one dataset to
    be clustered using the $k$-means algorithm.
}\label{fig:cluster}
\end{figure}

 In Figure~\ref{fig:Shared}, we illustrate the broad principles of our approach. 
 We factorize link and content over time at once.
 The matrix  $U$  represents link and content in a shared dimension
 and each snapshot of time will have a different  matrix $U$. These
 different values of the matrix $U$ provide insights at various
 temporal snapshots, and therefore they can be used for temporal
 community detection  as shown in Figure~\ref{fig:cluster}.

\begin{figure}[htbp]
\centering
\includegraphics[width=0.45\textwidth]{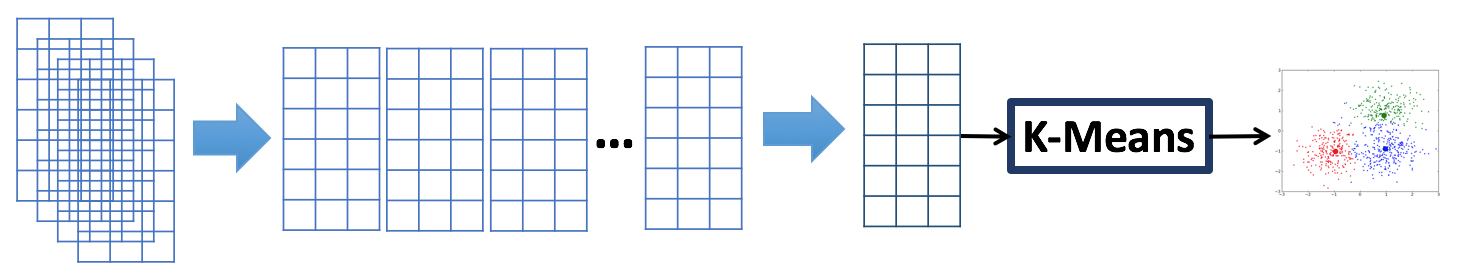}
\caption{%
    The different matrices corresponding to $U$ combine as a stream of data to
    predict future communities.}\label{fig:Prediction}
\end{figure}

Related to the problem of community detection is is that of community
prediction, in which one attempts to predict future communities from
previous snapshots of the network. It is notoriously difficult to predict
future clustering structures from a complex combination of data types such as
links and content. However, the matrix factorization methodology
provides a nice abstraction, because one can now use the multidimensional
representations created by sequences of matrices over different snapshots. The
basic idea is that we can consider each of the entries in the latent space
representation as a stream of evolving entries, which implicitly creates
a time-series that can be used to
 predict future communities, as illustrated in Figure~\ref{fig:Prediction}.

In this work, we present \emph{Chimera}, a method that uses link and content from networks over time to detect and
predict community structure. To the best of our knowledge, there is no work
addressing these three aspects \emph{simultaneously} for both detection and
prediction of communities. The main contributions of this paper are:

\begin{itemize}
\item An efficient algorithm based on shared matrix factorization that
      uses link and content over time; the uniform nature of the
      embedding allows the use of any traditional clustering
      algorithm on the corresponding representation.
\item A method for predicting future communities from embeddings over
      snapshots.
\end{itemize}

\section{Related Work}\label{sec:back}

In this section, we review the existing work for community detection using link
analysis, content analysis, temporal analysis and their combination. Since these
methods are often proposed in different contexts (sometimes even by different
communities), we will organize these methods into separate sections.

\paragraph{Topological Community Detection} These methods are based mainly on
links among nodes. The idea is to minimize the number of edges across nodes
belonging to different communities. Thus, the nodes inside the community should
have a higher density (number of edges) with other nodes inside the community
than with nodes outside the community.  There are several ways of defining and
quantifying communities based on their topology,
modularity~\cite{louvain}, conductance~\cite{Leskovec2008},
betweeness~\cite{girvan2002community}, and spectral partition~\cite{Barnes1982}.
More can be found in~\textcite{fortunato2010community}.

\paragraph{Content-Centric Community Detection} Topic modeling is a common
approach for content analysis and is often used for clustering, in addition to
dimensionality reduction.  PLSA-PHITS~\cite{hofman2008bayesian} and
LDA~\cite{Cohn:2000} are the most traditional methods for content analysis, but
they are susceptible to words that appear very few times. Extended methods that
are more reliable are Link-PLSA-LDA~\cite{Nallapati:2008} and
Community-User-Topic model~\cite{Zhou:2006}. In most cases, the combination of
link and content provides insights that are missing with the use of a single
modality.

\paragraph{Link and Temporal Community Detection} A few authors address the
problem of temporal community detection that aims to identify how communities
emerge, grow, combine, and decay over
time~\cite{lin2008facetnet,kawadia2012sequential},
\cite{chakrabarti2006evolutionary}, \cite{kim2009particle},
\textcite{tang2011dynamic} use temporal Dirichlet processes to detect
communities and track their evolution.  \textcite{chen2013detecting} tackle the
problem of overlapping temporal communities. \textcite{bazzi2016community}
propose the detection of communities in temporal networks represented as
multilayer networks. \textcite{Pietilanen:2012} identify clusters of nodes that
are frequently connected for long periods of time, and such sets of nodes are
referred to as temporal communities. \textcite{He201587} propose an algorithm
for dynamic community detection in temporal networks, which takes advantage of
community information at previous time steps. \textcite{Yu2017WSDM} present
a model-based matrix factorization for link prediction and also for community
prediction. However, their work uses only links for the prediction process.

\paragraph{Link and Content-Centric Community Detection} In recent years, some
approaches were developed to use link and content information for community
detection~\cite{Ruan:2013,Yang:2009,liu2015community,xu2014exploiting}. Among
them, probabilistic models have been applied to fuse content analysis and link
analysis in a unified framework. Examples include generative models that combine
a generative linkage model with a generative content-centric model through some
shared hidden variables~\cite{cohn2001missing,nallapati2008joint}.
A discriminative model is proposed by~\textcite{Yang:2009}, where a conditional
model for link analysis and a discriminative model for content analysis are
unified. In addition to probabilistic models, some approaches integrate the two
aspects from other directions. For instance, a similarity-based
method~\cite{zhou2009graph} adds virtual attribute nodes and edges to a network,
and computes the similarity based on the augmented network.
\textcite{gupta2010nonnegative} use matrix factorization to combine sources to
improving tagging.  It is evident that none of the aforementioned works combine
all the three factors of link, content, and temporal information within
a unified framework; caused in part by the fact that these modalities interact
with one another in complex ways. Therefore, the use of latent factors is
a particularly convenient way to achieve this goal.

\paragraph{Community Prediction} There has been a growing interest in the
dynamics of communities in evolving social networks, with recent studies
addressing the problem of building a predictive model for community detection.
Most of the community prediction techniques described in these works are about
community evolution prediction that aim to predict events such as growth,
survival, shrinkage, splits and
merges~\cite{Ngonmang2010,Bringmann2010,Ngonmang2013,Saganowski2013,Takaffoli2014,Sharma2015,Saganowski20152}.
In \cite{Ilhan2013,Ilhan2015} the authors use ARIMA models to predict community events in a network
without using any previous community detection method. \textcite{ILHAN2016}
propose to use a small number of features to predict community events.
\textcite{Pavlopoulou2017} employ several structural and temporal features to
represent communities and improve community evolution prediction.

\medskip

The community prediction addressed in our work can predict not only community evolution but also a more accurate prediction about each node of the network, in which community the node will be and if its community will change or not. We do so by using topological characteristics and also content associated with nodes.

\section{Problem Definition}\label{sec:problem}

We assume we have $T$ graphs $G_1 \ldots G_T$ that form a time-series.  The
graphs are defined over a fixed set of nodes $\mathcal{N}$ of cardinality $n$.
In each timestamp, a different set of edges may exist over time. For example, in
the case of a co-authorship network, the node set may correspond to the authors
in the network, and the graph $G_t$ might correspond to the co-author relations
among them in the $t$th year. These co-author relations are denoted by the $n
\times n$ adjacency matrix $A_t$.  Note that the entries in $A_t$ need not be
binary, but might contain arbitrary weights. For example, in a co-authorship
network, the entries might correspond to the number of publications between
a pair of authors.  For undirected graphs, the adjacency matrix $A_t$ is
symmetric, while in directed graphs the adjacency matrix is asymmetric. Our
approach can handle both settings.  Hence, the graph $G_t$ is denoted by the
pair $G_t=(\mathcal{N}, A_t)$.

We assume that for each timestamp $t$, we have an $n \times d$ content matrix
$C_t$.  $C_t$ contains one row for each node, and each row contains $d$
attribute values representing the content for that node at the $t$th timestamp.
For example, in the case of the co-authorship network, $d$ might correspond to
the lexicon size, and each row might contain the word frequencies of various
keywords in the titles. Therefore, one can fully represent the content and
structural pair at the $t$th timestamp with the triplet $(\mathcal{N}, A_t,
C_t)$.

In this paper, we study the problem of content-centric community detection in
networks. We study two problems: temporal community {\em detection}, and
community {\em prediction}. While the problem of temporal community detection
has been studied in the literature, as presented in Section~\ref{sec:back}, the
problem of community prediction, as defined in this work, has not been studied
to any significant extent.  We define these problems as follows.

\begin{definition}[Temporal Community Detection]
    Given a sequence of snapshots of graphs $G_1 \ldots G_T$, with $n \times n$
    adjacency matrices $A_1 \ldots A_T$, and $n \times d$ content matrices $C_1
    \ldots C_T$, create a clustering of the nodes into $k$ partitions  at each
    timestamp $t \leq T$.
\end{definition}

The clustering of the nodes at each timestamp $t$ may use only the graph
snapshots up to and including time $t$.  Furthermore, the clusters in
successive timestamps should be temporally related to one another. Such
a clustering provides better insights about the evolution of the graph.
In this sense, the clustering of the nodes for each
timestamp will be somewhat different from what is obtained using an
independent clustering of the nodes at each timestamp.

\begin{definition}[Temporal Community Prediction]
Given a sequence of snapshots of graphs $G_1 \ldots G_T$ with $n
\times n$ adjacency matrices $A_1 \ldots A_T$, and $n \times d$
content matrices $C_1 \ldots C_T$, {\bf \emph{predict}} the clustering of
the nodes into $k$ partitions at {\bf \emph{future}} timestamp $T+r$.
\end{definition}

The community prediction problem attempts to predict the communities at a {\em
future timestamp}, before the structure of the network is known.  To the best of
our knowledge, this problem is new, and it has not been investigated elsewhere
in the literature.  Note that the temporal community prediction problem is more
challenging than temporal community detection, because it requires us to predict
the community structure of the nodes {\em without any knowledge of the adjacency
matrix} at that timestamp.

Temporal prediction is generally a much harder problem in the
structural domain of networks as compared to the multidimensional
setting. In the multidimensional domain, one can use numerous
time-series models such as the  auto-regressive (AR) model to
predict future trends. However, in the  structural domain, it is far
more challenging to make such predictions.

\section{Mathematical Model}\label{sec:model}

In this section, we discuss the optimization model for
converting the temporal sequences of graphs and content to a
multidimensional time-series.  To achieve this goal, we use a
non-negative matrix factorization framework.  Although the
non-negativity is not essential, one advantage is that it leads to a
more interpretable analysis. Consider a setting in which the rank of
the factorization is denoted by $k$.  The basic idea is to use three
sets of latent factor matrices in a {\em shared} factorization
process,  which is able to combine content and  structure in a
holistic way:

\begin{enumerate}
\item The matrix $U_t$ is an $n \times k$ matrix, which is specific
to each timestamp $t$.  Each row of the matrix $U_t$ describes the
$k$-dimensional latent factors of  the corresponding node at time
stamp $t$, while taking into account {\em both} the structural and
content information.
\item The matrix $V$ is an $n \times k$ matrix, which is global to
all timestamps. Each row of the matrix $V$ describes the
$k$-dimensional latent factors of the corresponding node over all
time stamps, based on {\em only} the structural information.
\item The matrix $W$ is an $d \times k$ matrix, which is global to
all timestamps. Each row of the matrix $W$ describes the
$k$-dimensional latent factors of  one of the $d$ keywords over all
time stamps, based on {\em only} the content information.
\end{enumerate}

The matrices $U_1 \ldots U_T$ are more informative than the other matrices,
because they contain latent information specific to the content and structure,
and they are also specific to each timestamp. However, the matrices $V$ and $W$
are global, and they contain {\em only} information corresponding to the
structure and the content in the nodes, respectively. This is a setting that is
particularly suitable to {\em shared} matrix factorization, where the matrices
$U_1 \ldots U_T$ are shared between the factorization of the adjacency and
content matrices. 

Therefore, we would like to approximately factorize the adjacency matrices $A_1
\ldots A_T$ as $A_t \approx U_t V^T$, for all $t \in \{ 1 \ldots T \}$.
Similarly, we would like to approximately factorize the content matrices $C_1
\ldots C_T$ as $C_t \approx U_t W^T$. 
With this setting, we propose the following optimization problem:
\begin{equation}
\begin{aligned}
    J&= \sum_{t=1}^T \left\|A_t - U_t V^T\right\|^2
      + \beta \sum_{t=1}^T \left\| C_t - U_t W^T\right\|^2 \\
      &+ \lambda_1 \Omega(U_t, V, W)\,\text{.}
\end{aligned}
\end{equation}
Where $\beta$ is a balancing parameter, $\lambda_1$ is the
regularization parameter, and $\Omega(U_t, V, W)$ is a
regularization term to avoid overfitting. The notation $\|\cdot\|^2$
denotes the Frobenius norm, which is the sum of the squares of the
entries in the matrix. The regularization term is defined as
\begin{equation}
\Omega(U_t, V, W)= \|V\|^2 + \|W\|^2 + \sum_{t=1}^T \left\|U_t\right\|^2\,\text{.}
\end{equation}
We would also like to ensure that the embeddings between successive timestamps
do not change suddenly because of random variations. For example, an author
might publish together with a pair of authors every year, but might not be
publishing in a particular year because of random variations. To ensure that the
predicted values do not change suddenly, we add a temporal regularization term:
\begin{equation}
\Omega_2(U_1 \ldots U_T)= \sum_{t=1}^{T-1} \| U_{t+1} - U_{t}\|^2
\end{equation}
This additional regularization term ensures the variables in any pair of
successive years do not change suddenly. The additional regularization term is
added to the objective function, after multiplying it with $\lambda_2$.
The enhanced objective function is defined as
\begin{equation}
\begin{aligned}
J &= \sum_{t=1}^T \|A_t - U_t V^T\|^2 + \beta \sum_{t=1}^T \| C_t -
U_t W^T\|^2 +\\
&+ \lambda_1 \left( \|V\|^2 + \|W\|^2 + \sum_{t=1}^T \|U_t\|^2\right) +
  \lambda_2 \sum_{t=1}^{T-1} \| U_{t+1} - U_{t}\|^2\,\text{.}
\end{aligned}\label{eq:enhanced-model}
\end{equation}

In order to ensure a more interpretable solution, we impose non-negativity
constraints on the factor matrices
\begin{equation}
U_t \geq 0, V \geq 0, W \geq 0\,\text{.}\label{eq:non-negative}
\end{equation}

One challenge with this optimization model is that it can become very large.
The main size of the optimization model is a result of the adjacency matrix. The
content matrix is often manageable, because one can often reduce the
keyword-lexicon in many real settings. However, the adjacency matrix scales with
the square of the number of nodes, which can be onerous in real settings. An
important observation here is that the adjacency matrix is sparse, and most of
its values are zeros. Therefore, one can often use sampling on the zero entries
of the  adjacency matrix in order to reduce the complexity of the problem.  This
also has a beneficial effect of ensuring that the solution is not dominated by
the zeros in the matrix.

\subsection{Solving the Optimization Model} \label{sec:solve}

In this section, we discuss a gradient-descent approach for
solving the optimization model. The basic idea is to compute the
gradient of $J$ with respect to the various parameters. Note that
$U_t V^T$ can be seen as the ``prediction'' of the value of
$A_t$. Obviously, this predicted value may not be the same as
the observed entries in the adjacency matrices. Similarly, while the
product $U_t W^T$ predicts $C_t$, the predicted values
may be different from the observed values.  The gradient descent
steps are dependent on the errors of the prediction.  Therefore, we
define the error for the structural and content-centric entries as
$\Delta^A_t = A_t - U_tV^T$ and $\Delta^C_t = C_t - U_tW^T$.
Also let $\Delta^U_t = U_t - U_{t+1}$, with $\Delta^U_T = 0$, since the
difference is not defined at this boundary value.

Our goal is to compute the partial derivative of $J$ with respect to the various
optimization variables, and then use it to construct the gradient-descent steps.
By computing the partial derivatives of \eqref{eq:enhanced-model} with respect
to each of the decision variables, we obtain
\begin{align}
    \frac{\partial J}{\partial U_t} &= 2\lambda_1 U_t - 2\left(\Delta^A_tV + \beta \Delta^C_tW\right) +
    2\lambda_2 \Delta^U_t\,,\label{eq:partial-u} \\
    \frac{\partial J}{\partial V_t} &= 2\lambda_1 V_t - 2\sum_{t=1}^T \left[\Delta^A_t\right] U_t \label{eq:partial-v}\\
    \frac{\partial J}{\partial W_t} &= 2\lambda_1 W_t - 2\beta\sum_{t=1}^T \left[\Delta^C_t\right] U_t\,.\label{eq:partial-w}
\end{align}

The gradient-descent  steps use these partial derivatives for the
updates. The gradient-descent steps may be written as
\begin{align}
    U_t &\leftarrow U_t - \alpha \frac{\partial J}{\partial U_t} \ \forall t\,,\label{eq:gd-updates-u}\\
    V &\leftarrow V_t - \alpha \frac{\partial J}{\partial V} \,,\label{eq:gd-updates-v}\\
    W &\leftarrow W_t - \alpha \frac{\partial J}{\partial W}\,\label{eq:gd-updates-w}.
\end{align}
Here, $\alpha>0$ is the step-size, which is a small value, such as
0.01. The matrices $U_t$, $V$, and $W$ are initialized to non-negative values in
$(0, 1)$, and the updates~(\ref{eq:gd-updates-u}--\ref{eq:gd-updates-w}) are
performed until convergence or until a pre-specified number of iterations is
performed.  Non-negativity constraints are enforced by setting an entry in these
matrices to zero whenever it becomes negative due to the updates.

$\Delta^A_t$ is a sparse matrix, and should be stored using sparse data
structures. As a practical matter, it makes sense to first compute those entries
in $\Delta^A_t$ that correspond to non-zero entries in $A$, and then store those
entries using a sparse matrix data structure. This is because a $n \times n$
matrix may be too large to hold using a non-sparse representation.

Combining equations~(\ref{eq:partial-u}--\ref{eq:gd-updates-w}), we obtain the
following update rule:
\begin{equation}
\begin{aligned}
U_t &\leftarrow U_t( 1 - 2 \alpha \lambda_1) + 2 \alpha \Delta^A_t V
+
2 \alpha \beta \Delta^C_t W + 2 \lambda_2 \Delta^U_{t}\\
V &\leftarrow V (1 - 2 \alpha  \lambda_1) + 2 \alpha
\sum_{t=1}^T [\Delta^A_t]^T U_t\\
W &\leftarrow W (1 - 2 \alpha  \lambda_1) + 2 \alpha \beta
\sum_{t=1}^T [\Delta^C_t]^T U_t\,.\\
\end{aligned}\label{eq:derivatives}
\end{equation}

The  set of  updates above  are typically performed
``simultaneously'' so that the entries in $U_t$, $V$ and $W$  (on
the right-hand side) are fixed to their values in the previous
iteration during a particular block of updates. Only after the new
values of $U_t$, $V$, and $W$ have been  computed (using temporary
variables), can they be used in the right-hand side in the next
iteration.

\subsection{Complexity Analysis}\label{sec:complexity}

With the algorithm fully specified, we can now analyze its asymptotic
complexity. Per gradient descent iteration, the computational cost of the
algorithm is the sum of (i) the complexity of evaluating the objective
function~\eqref{eq:enhanced-model} and (ii) the complexity of the
update step~\eqref{eq:derivatives}. Recall from section~\ref{sec:model} that
$A_t$, $C_t$, $U_t$, $V$, and $W$ have dimensions $n\times n$, $n\times d$,
$n\times k$, $n\times k$, and $d\times k$, respectively. Since matrix
factorization reduces the dimensions of the data, we can safely assume $n \gg
k$ and that $d \gg k$.

Assuming the basic matrix multiplication algorithm is used, the complexity of
multiplying matrices of dimensions $m\times p$ and $p\times n$ is $O(mnp)$.
Therefore, the complexity of computing $\|U_tV^T\|^2 = O(n^2k) + O(n^2)$, since
the norm can be computed by iterating over all elements of the matrix, squaring
and summing them. Hence, the complexity of evaluating the objective
function~\eqref{eq:enhanced-model} is
\begin{equation*}
    \begin{aligned}
        J =&\, T\left[O(n^2k) + O(n^2) + O(dkn) + O(dn) + O(kn) + O(kn)\right] \\
           & + \, O(kn) + O(dk) \\
          =&\, O(\max(n^2k, dkn))\,.
    \end{aligned}
\end{equation*}

To obtain the asymptotic complexity of the updates, note that
$\Delta^A_t=A_t-U_tV^T$, and $\Delta^C_t=C_t-U_tW^T$. Hence, $\Delta^A_tV
= O(n^2k)$, $[\Delta^A_t]^TU_t=O(n^2k)$, $\Delta^C_tW = O(dkn)$, and
$[\Delta^C_t]^TU_t=O(dkn)$. Therefore, the asymptotic complexity of the
gradient descent update is
   $ T [O(kn) + O(n^2k) + O(dkn)] = O(\max(n^2k, dkn))\,$.

\section{Applications to clustering}

\subsection{Temporal Community Detection}\label{sec:temporal-community-detection}

The learned factor matrices can be used for temporal community
detection. In this context, the matrix $U_t$ is very helpful in
determining the communities at time $t$, because it accounts for
structure, content, and smoothness constraints. The
overall approach is:

\begin{enumerate}
    \item Extract the $n \cdot T$ rows from $U_1 \ldots U_T$, so that each of
        the $n \cdot T$ rows is associated with a timestamp from $\{ 1 \ldots
        T \}$. This timestamp will be used in step 3 of the algorithm.
    \item Cluster the $n \cdot T$ rows into $k$ clusters $\mathcal{C}_1 \ldots
        \mathcal{C}_k$ using a $k$-means clustering algorithm.
    \item Partition each $\mathcal{C}_i$ into its $T$ different timestamped
        clusters $\mathcal{C}^1_i \ldots \mathcal{C}^T_i$, depending on the
        timestamp of the corresponding rows.
\end{enumerate}

In most cases, the clusters will be such that the $T$ different
avatars of the $i$th row in $U_1 \ldots U_T$ will belong to the same
cluster. However, in some cases, rows may drift from one cluster to
the other. Furthermore, some clusters may shrink with time, whereas
others may increase with time. All these aspects provide interesting
insights about the community structure in the network. Even though
the data is clustered into $k$ groups, it is often possible for one
or more timestamps to contain clusters without any members. This is
likely when the number of clusters expands or shrinks with time.

\subsection{Temporal Community Prediction}

This approach can also be naturally used for community prediction.  The basic
idea here is to treat $U_1 \ldots U_T$ as a time-series of matrices, and
predict how the weights evolve with time. The overall approach is as follows:

\begin{enumerate}
    \item For each $(i, j)$ of the non-zero entries of matrix $A$,
          represent the time series $\mathcal{T}_{ij}$.
    \item Use an autoregressive model on $\mathcal{T}_{ij}$ to predict
          $u_{ij}^{t+r}$ for each $(i, j)$ of the non-zero entries. Set all
          other entries in $U_{t+r}$ to 0.
    \item Perform node clustering on the rows of $U_{t+r}$ to create
          the predicted node clusters at time $(t+r)$. This provides the
          predicted communities at a future timestamp.
\end{enumerate}
Thus, \emph{Chimera} can provide not only the communities in the
current timestamp, but also the communities in a future
timestamp. 

\begin{table*}
\small
 \begin{center}
    \caption{
        Jaccard (J) and Purity (P) of the {\em Synthetic 1}  and {\em Synthetic 2} dataset from all timestamps
        and methods. \emph{Chimera} outperforms baseline methods in almost every year.
    }\label{tab:sin2}
    \begin{tabular}{lcccccc|cccccc}
        \toprule
        & \multicolumn{6}{c}{Synthetic 1} &  \multicolumn{6}{c}{Synthetic 2} \\
         \cmidrule{2-13}
        &                 \multicolumn{2}{c}{1}          & \multicolumn{2}{c}{2} &    \multicolumn{2}{c}{3}   &  \multicolumn{2}{c}{1}          & \multicolumn{2}{c}{2} &    \multicolumn{2}{c}{3}  \\
        \cmidrule{2-13}
        Algorithm  & J      & P  & J     & P  & J       & P  & J      & P  & J     & P  & J       & P         \\
        \midrule
        Louvain    & 0.4   & 1  & 0.2  & 1  & 0.4   & 1  & 0.2  & 1   & 0.2   & 1    & 0.2      & 1   \\
        GibbsLDA++        & 0.542    & 0.714   & 0.267   & 0.463   & 0.399     & 0.611   & 0.279  & 0.473    & 0.533     & 0.657   & 0.326  & 0.575 \\
        CPIP-PI    & 0.909    & 1       & 1       & 1       & 0.866     & 0.917  & 1      & 1        & 0.999     & 0.997  & 0.999  & 0.995 \\
        DCTN       & 0.4        & 1       & 0.2       & 1       & 0.2         & 1      & 0.2      & 1        & 0.2     & 1    & 0.2      & 1       \\
        Chimera        & 1    & 1   & 1   & 1   & 1    & 1   &  0.999 & 0.994    & 0.998     & 0.992   & 0.997  & 0.990  \\
        \bottomrule
    \end{tabular}
    \end{center}
\end{table*}

\section{Experiments}\label{sec:experiments}

This section describes the experimental results of the approach.  We describe
the datasets, evaluation methodology, and the results obtained.

A key point in choosing a dataset to evaluate algorithms such as \emph{Chimera}
is that there must be co-evolving interactions between network and content.  In
order to check our model's consistency, and to have a fair comparison with other
state-of-the-art algorithms, we generated a couple of synthetic dataset. 

\paragraph{Synthetic dataset} The synthetic dataset was generated in the
following way: first, we create the matrix $A_1$ with 5 groups. Then, we follow
a randomized approach to rewire edges. According to some probability, we connect
edges from one group to another.  In this dataset, all link matrices ($A$) have
5,000 nodes and 20,000 edges. For the content matrices ($C$), we generate five
groups of five words. As in the link case, we have a probability of a word being
in more than one group. Due to the nature of its construction, all content
matrices have 25 words.  For transitioning between timestamps, we have another
probability that defines whether a node changes group or not. The transitions
are constrained to be at most 10\% of the nodes.  We generated
3 timestamps for each synthetic dataset.  The rewire probabilities $1-p$ used in
each synthetic dataset were $p=0.75$ (Synthetic 1) and $p=0.55$ (Synthetic 2).

\paragraph{Real dataset} We used the arXiv
API\footnote{\texttt{\url{https://arxiv.org/help/api/index}}} to download information
about preprints submitted to the arXiv system. We extracted information about
7107 authors during a period of five years (from 2013 to 2017). We used the
papers' titles and abstracts to build the author-content network with
10256 words, and we selected words with more than 25 occurrences after removal
of stop words and stemming. Since every preprint submitted to the arXiv
has a category, we used the category information as a group label. We selected
10 classes: cs.IT, cs.LG, cs.DS, cs.CV, cs.SI, cs.AI, cs.NI, cs, math, and stat.
Authors were added to the set of authors if they published for at least three
years in the five-year period we consider. In years without publications, we
assume authors belong to the temporally-closest category.

There are several metrics for evaluating cluster quality. We use two well-known
supervised metrics: the Jaccard index and cluster purity.
Cluster purity~\cite{Manning:2008} measures the quality of the communities by
examining the dominant class in a given cluster. It ranges from 0 to 1, with
higher purity values indicating better clustering performance. 

We compared our approach  with state-of-the-art algorithms in four categories:
Content-only, Link-only, Temporal-Link-only and Link-Content-only.  By following
this approach, we are also able to isolate the specific effects of using data in
different modalities.

\textbf{Content-only method.} We use GibbsLDA++ as a  baseline for the
content-only method. As input for this method, we considered that a document
consists of the words used in the title and abstract of a paper.

\textbf{Link-only method.} For link we use the
Louvain~\cite{louvain} method for community detection. 
\textbf{Temporal-Link-only method.} For temporal link-only method we
used the work presented by~\textcite{He201587}, which we refer to
as DCTN. 

\textbf{Combination of Link and Content\footnote{Code from authors
obtained from
\texttt{\url{https://github.com/LiyuanLucasLiu/Content-Propagation}}.}} For link
and content combination, we used the work presented
by~\textcite{liu2015community}, with algorithms CPRW-PI, CPIP-PI, CPRW-SI,
CPIP-SI. Since all them perform very similarly and we have a space constraint we
will report only the results obtained with CPIP-PI.

\subsection{Evaluation Results}

\begin{table*}[ht!]
\small
\begin{center}
\caption{
    Purity (P) and Jaccard (J) Index obtained in the \emph{arXiv} dataset for all
    years and methods. \emph{Chimera} outperforms baseline methods in almost
    every year.
}\label{tab:results}
    \begin{tabular}{lcccccccccc}  
        \toprule
                        & \multicolumn{2}{c}{2013}          & \multicolumn{2}{c}{2014}             & \multicolumn{2}{c}{2015}          & \multicolumn{2}{c}{2016}          & \multicolumn{2}{c}{2017}  \\
        \cmidrule{2-11}
        Algorithm       & J         & P          & J          & P            & J         & P          & J         & P          & J         & P  \\
        \midrule
        Louvain         & 0               & 0.041          & 0                & 0.062            & 0               & 0.073          & 0               & 0.100          & 0               & 0.086  \\ 
        GibbsLDA++      & 0.087          & 0.373          & 0.080           & 0.394            & 0.182   & 0.387         & \textbf{0.166}          & 0.399          & 0.168          & 0.389  \\
        CPIP-PI         & \textbf{0.096}          & \textbf{0.523}          & 0.097           & 0.518            & 0.149          & 0.412          & 0.090          & 0.365          & 0.105          & 0.361  \\ 
        DCTN            & 0               & 0.039          & 0                & 0.052         & 0               & 0.069          & 0               & 0.077          & 0               & 0.085  \\ 
        \emph{Chimera}      & 0.078          & 0.456          & \textbf{0.261}     & \textbf{0.601}    & \textbf{0.281}  & \textbf{0.610}  & 0.105   & \textbf{0.573}  & \textbf{0.291}   & \textbf{0.628}  \\ 
        \bottomrule
    \end{tabular}
\end{center}
\end{table*}

In this section, we present the results of our experiments.

The Louvain and DCTN methods are based on link structure and do not
allow fixed numbers of clusters. They use topological structure to
find the number of communities.
All methods in the baseline were used in their default
configuration.

First, we present the results with synthetic data we generated (Synthetic 1 and
Synthetic 2) in Table~\ref{tab:sin2}. In synthetic datasets we use
$\alpha=0.00001$,  $\beta=1000$,  $\lambda=0.1$ and $\lambda_2=0.0001$ with
$k=5$ and 1000 steps. 

The only methods that are able to find the clusters in all datasets are CPIP-PI
and \emph{Chimera}, both using content and link information.  In the synthetic
data the changes between timestamps were small. Thus, CPIP-PI and
\emph{Chimera} performed similarly.  However, \emph{Chimera} displayed almost
perfect performance in all datasets and timestamps. 
Louvain and DCTN, which use only link information, were not able to find the
clusters. Despite the purity of 1, they cluster all the data into only one
cluster.  DCTN finds clusters only for the two first timestamps of 
synthetic 2, obtaining 3 and 4 clusters respectively. Louvain found 3 clusters in
timestamps 1 and 3 of synthetic 2. 

Table \ref{tab:results} presents the  Jaccard and Purity metrics over all
methods for the real dataset {\em arXiv}.  In {\em arXiv}, the Louvain method
found 3636, 2679, 2006, 1800 and 2190 communities respectively for each year.
CDTN, which is based on Louvain has a very similar result with 3636, 2656, 1829,
1500 and 1791 communities respectively for each year.  Since they are methods
based on link, they consider specially disconnected nodes as isolated
communities.  Methods that combine link and content use content to
aggregate such nodes in a community.  Also, as we can note in
Table~\ref{tab:results}, our method can learn with time and improve its
results in the following years.  GibbsLDA++ presents a nice performance
because the content was much more stable and had more quality over the
years than the link information. This is another reason to combine various
sources to achieve better performance. 

To tune the hyperparameters of \emph{Chimera}, we used Bayesian
Optimization~\cite{bergstra2013making,shahriari2016taking} to perform a search
in the hyperparameter space. Bayesian Optimization is the appropriate technique
in this setting, because minimizing the model loss~\eqref{eq:enhanced-model}
does not necessarily translate into better performance. We defined an objective
function that minimizes the mean silhouette
coefficient~\cite{rousseeuw1987silhouettes} of the labels assigned by
\emph{Chimera}, as described in section~\ref{sec:temporal-community-detection}.
We used Bayesian Optimization to determine the number of clusters as well.
With this approach, the optimization process is completely unsupervised and,
although we have access to the true labels, they were not used during
optimization, a situation closer to reality. With Bayesian Optimization, our
model was able to learn that the actual number of clusters was in the order of
10.  The full set of hyperparameters and their ranges are shown in
Table~\ref{tab:hyperparameters}, with best results shown in bold face.

\begin{table}[t]
    \caption{
        Hyperparameters used for tuning \emph{Chimera} with Bayesian Optimization.
        Elements in bold indicate the best parameter for that hyperparameter.
        The set of all elements in bold defines the hyperparameters used for
        training the model.
    }\label{tab:hyperparameters}
    \begin{center}
    \begin{tabular}{lr}
        \toprule
        Hyperparameter     & Values \\
        \midrule
        $\alpha$           & \{$0.01, \mathbf{0.1}$\}   \\
        $\beta$            & \{$0.1, 0.25, 0.5, 0.75, \mathbf{0.9}$\}   \\
        $\lambda_1$        & \{$\mathbf{1\times10^{-5}}, 1\times10^{-6}$\}   \\
        $\lambda_2$        & \{$1\times10^{-4}, \mathbf{1\times10^{-5}}$\}   \\
        $K$                & \{$\mathbf{10}, 20, 30, 40, 50$\}   \\
        Clusters           & \{$2, 4, 8, \mathbf{10}, 16, 18, 32$\}   \\
        \bottomrule
    \end{tabular}
    \end{center}
\end{table} 

\begin{table}
\caption{
    The Jaccard index and Purity of {\em arXiv} for prediction. In the ``Original
    U's'' row, we used the original matrices to make the prediction in 2015,
    2016 and 2017. Whereas in the ``Predicted U's'' row, we used the output of
    \emph{Chimera} to make the predictions. Hence, for 2016 we used the
    prediction for 2015, and for 2017 we used the predictions of both 2015
    and 2016.
} \label{tab:resultsPred}
\begin{center}
    \hspace*{-2.5mm}\begin{tabular}{lcccccc}
    \toprule
    & \multicolumn{2}{c}{2015} & \multicolumn{2}{c}{2016}  & \multicolumn{2}{c}{2017} \\
    \midrule
    & J & P & J & P & J & P  \\
    \midrule
    Original U's & 0.0709 & 0.5180  & 0.2273 & 0.4395 &  0.1145 & 0.3766 \\
    Predicted U's  & &   & 0.0589 & 0.5177 &  0.0765 & 0.4981 \\
    \bottomrule
\end{tabular}
\end{center}
\end{table} 

In Table \ref{tab:resultsPred} we show our results for prediction. Here, we will
not compare our results with other methods that estimate or evaluate the size of
each community. The idea here is to predict in which community an author will be
in the future.  One advantage of our method is that we can augment our time
series with our predictions.  Clearly, doing so will add noise to further
predictions, but the results presented are very similar to the ones present in
the original dataset.  \emph{Chimera} is the only one that allows us to do that
kind of analysis in an easy way, since the embeddings create multidimensional
representations of the nodes in the graph. 

\subsection{Performance evaluation}

In section~\ref{sec:complexity}, we claimed our algorithm is $O(\max(n^2kT,
dknT)$. In this section we evaluate the performance of the algorithm for
datasets generated following the rules of generation of
section~\ref{sec:experiments}.

\begin{figure}
\begin{center}
    \includegraphics[width=\linewidth]{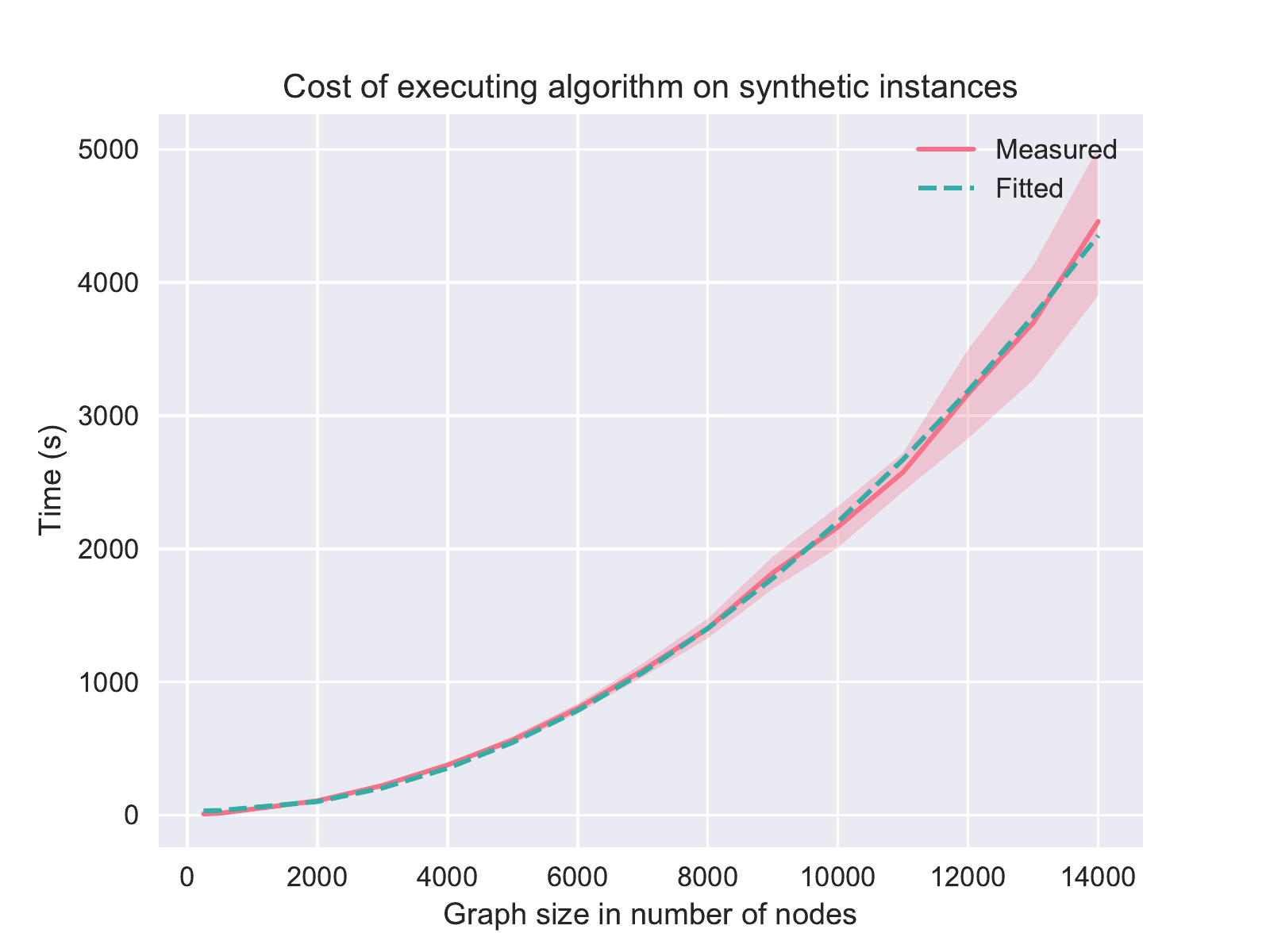}
    \caption{%
        Performance of the proposed algorithm in synthetic datasets.
    }\label{fig:times}
\end{center}
\end{figure}

We generated 15 datasets of increasing sizes (with $n$ ranging from 250 to
14,000). Since $d \ll n$ in these datasets, we expect \emph{Chimera}'s asymptotic
complexity to be $O(n^2k)$. To verify this, we measure the time it took to
execute 1000 iterations of $\emph{Chimera}$ with $T=3$, $k = 2$, $\alpha=0.001$,
$\beta=0.001$, $\lambda=0.005$, $\lambda_2=0.001$. Being $T$ and $k$ small
integers, it is expected the $n^2$ factor will dominate the growth of the
algorithm.  To know whether that is the case, we also fit the data to a degree
two polynomial that minimizes the squared error. The obtained data is
summarized in Figure~\ref{fig:times}. As can be seen from the figure, there is
a good fit between the measured data and the fitted polynomial, indicating the
order of growth is quadratic for datasets similar to the ones presented.

\section{Conclusions} \label{sec:conc}

In this work, we presented \emph{Chimera} a novel shared factorization overtime model that can
simultaneously take the link, content, and temporal information of networks into
account improving over the state-of-the-art approaches for community detection. 
Our approach model and solve in efficient time the problem of combining link,
content and temporal analysis for community detection and prediction in network
data. Our method extracts the latent semantic structure of the network in
multidimensional form, but in a way that takes into account the temporal
continuity of the embeddings. Such approach greatly simplifies temporal
analysis of the underlying network by using the embedding as a surrogate.
A consequence of this simplification is that it is also possible to use this
temporal sequence of embeddings to predict future communities with good results.  
The experimental results illustrate the effectiveness of \emph{Chimera},
since it outperforms the baseline methods. Our experiments also show that the
prediction is efficient in using embeddings to predict near future communities,
which opens a vast array of  new possibilities for exploration.

\section{Acknowlegments}

Charu C. Aggarwal's research was sponsored by the Army Research Laboratory and was accomplished under Cooperative Agreement Number W911NF-09-2-0053. The views and conclusions contained in this document are those of the authors and should not be interpreted as representing the official policies, either expressed or implied, of the Army Research Laboratory or the U.S. Government. The U.S. Government is authorized to reproduce and distribute reprints for Government purposes notwithstanding any copyright notation here on.

\balance
\printbibliography

@INPROCEEDINGS{Takaffoli2014,
    author={M. Takaffoli and R. Rabbany and O. R. Zaïane},
    booktitle={2014 IEEE/ACM International Conference on Advances in Social Networks Analysis and Mining (ASONAM 2014)},
    title={Community evolution prediction in dynamic social networks},
    year={2014},
    volume={},
    number={},
    pages={9-16},
    doi={10.1109/ASONAM.2014.6921553},
    ISSN={},
}

@INPROCEEDINGS{Pavlopoulou2017,
author={M. E. G. Pavlopoulou and G. Tzortzis and D. Vogiatzis and G. Paliouras},
booktitle={2017 12th International Workshop on Semantic and Social Media Adaptation and Personalization (SMAP)},
title={Predicting the evolution of communities in social networks using structural and temporal features},
year={2017},
volume={},
number={},
pages={40-45},
doi={10.1109/SMAP.2017.8022665},
ISSN={},
}

@INPROCEEDINGS{Ngonmang2013,
author={B. Ngonmang and E. Viennet},
booktitle={2013 International Conference on Signal-Image Technology Internet-Based Systems},
title={Toward Community Dynamic through Interactions Prediction in Complex Networks},
year={2013},
volume={},
number={},
pages={462-469},
doi={10.1109/SITIS.2013.81},
ISSN={},
}

@INPROCEEDINGS{Ngonmang2010,
author={M. Jamali and G. Haffari and M. Ester},
booktitle={2010 IEEE International Conference on Data Mining Workshops},
title={Modeling the Temporal Dynamics of Social Rating Networks Using Bidirectional Effects of Social Relations and Rating Patterns},
year={2010},
volume={},
number={},
pages={344-351},
doi={10.1109/ICDMW.2010.103},
ISSN={2375-9232},
}

@INPROCEEDINGS{Saganowski20152,
author={S. Saganowski},
booktitle={2015 IEEE/ACM International Conference on Advances in Social Networks Analysis and Mining (ASONAM)},
title={Predicting community evolution in social networks},
year={2015},
volume={},
number={},
pages={924-925},
doi={10.1145/2808797.2809353},
ISSN={},
}

@INPROCEEDINGS{Sharma2015,
author={A. Sharma and R. Kuang and J. Srivastava and X. Feng and K. Singhal},
booktitle={2015 IEEE/ACM International Conference on Advances in Social Networks Analysis and Mining (ASONAM)},
title={Predicting small group accretion in social networks: A topology based incremental approach},
year={2015},
volume={},
number={},
pages={408-415},
doi={10.1145/2808797.2808914},
ISSN={},
}

@INPROCEEDINGS{Saganowski2013,
author={B. Gliwa and P. Bródka and A. Zygmunt and S. Saganowski and P. Kazienko and J. Koźlak},
booktitle={2013 IEEE/ACM International Conference on Advances in Social Networks Analysis and Mining (ASONAM 2013)},
title={Different approaches to community evolution prediction in blogosphere},
year={2013},
volume={},
number={},
pages={1291-1298},
doi={10.1145/2492517.2500231},
ISSN={},
}

@INPROCEEDINGS{Ilhan2013,
author={N. Ilhan and I. G. Ögüdücü},
booktitle={2013 12th International Conference on Machine Learning and Applications},
title={Community Event Prediction in Dynamic Social Networks},
year={2013},
volume={1},
number={},
pages={191-196},
doi={10.1109/ICMLA.2013.40},
ISSN={},
}

@ARTICLE{Bringmann2010,
author={B. Bringmann and M. Berlingerio and F. Bonchi and A. Gionis},
journal={IEEE Intelligent Systems},
title={Learning and Predicting the Evolution of Social Networks},
year={2010},
volume={25},
number={4},
pages={26-35},
doi={10.1109/MIS.2010.91},
ISSN={1541-1672},
}

@inproceedings{Ilhan2015,
 author = {\.{I}lhan, Nagehan and \"{O}\u{g}\"{u}d\"{u}c\"{u}, \c{S}ule G\"{u}nd\"{u}z},
 title = {Predicting Community Evolution Based on Time Series Modeling},
 booktitle = {Proceedings of the 2015 IEEE/ACM International Conference on Advances in Social Networks Analysis and Mining 2015},
 series = {ASONAM '15},
 year = {2015},
 isbn = {978-1-4503-3854-7},
 location = {Paris, France},
 pages = {1509--1516},
 numpages = {8},
 url = {http://doi.acm.org/10.1145/2808797.2808913},
 doi = {10.1145/2808797.2808913},
 acmid = {2808913},
 publisher = {ACM},
}

@article{ILHAN2016,
title = {Feature identification for predicting community evolution in dynamic social networks},
journal = {Engineering Applications of Artificial Intelligence},
volume = {55},
pages = {202 - 218},
year = {2016},
issn = {0952-1976},
doi = {https://doi.org/10.1016/j.engappai.2016.06.003},
url = {http://www.sciencedirect.com/science/article/pii/S0952197616301117},
author = {\.{I}lhan, Nagehan and \"{O}\u{g}\"{u}d\"{u}c\"{u}, \c{S}ule G\"{u}nd\"{u}z}
}

@inproceedings{gupta2010nonnegative,
  title={Nonnegative shared subspace learning and its application to social media retrieval},
  author={Gupta, Sunil Kumar and Phung, Dinh and Adams, Brett and Tran, Truyen and Venkatesh, Svetha},
  booktitle={Proceedings of the 16th ACM SIGKDD international conference on Knowledge discovery and data mining},
  pages={1169--1178},
  year={2010},
  organization={ACM}
}

@book{Manning:2008,
 author = {Manning, Christopher D. and Raghavan, Prabhakar and Sch\"{u}tze, Hinrich},
 title = {Introduction to Information Retrieval},
 year = {2008},
 isbn = {0521865719, 9780521865715},
 publisher = {Cambridge University Press},
 address = {New York, NY, USA},
}

@article{louvain,
  author={Vincent D Blondel and Jean-Loup Guillaume and Renaud Lambiotte and Etienne Lefebvre},
  title={Fast unfolding of communities in large networks},
  journal={Journal of Statistical Mechanics: Theory and Experiment},
  volume={2008},
  number={10},
  pages={P10008},
  year={2008}
}

@article{He201587,
title = "A fast algorithm for community detection in temporal network ",
journal = "Physica A: Statistical Mechanics and its Applications ",
volume = "429",
number = "",
pages = "87 - 94",
year = "2015",
note = "",
issn = "0378-4371",
doi = "http://dx.doi.org/10.1016/j.physa.2015.02.069",
url = "http://www.sciencedirect.com/science/article/pii/S0378437115001922",
author = "Jialin He and Duanbing Chen"
}

@article{hofman2008bayesian,
  title={Bayesian approach to network modularity},
  author={Hofman, Jake M and Wiggins, Chris H},
  journal={Physical review letters},
  volume={100},
  number={25},
  pages={258701},
  year={2008},
  publisher={APS}
}

@inproceedings{Zhou:2006,
 author = {Zhou, Ding and Manavoglu, Eren and Li, Jia and Giles, C. Lee and Zha, Hongyuan},
 title = {Probabilistic Models for Discovering e-Communities},
 booktitle = {Proceedings of the 15th International Conference on World Wide Web},
 series = {WWW '06},
 year = {2006},
 isbn = {1-59593-323-9},
 location = {Edinburgh, Scotland},
 pages = {173--182},
 numpages = {10},
 url = {http://doi.acm.org/10.1145/1135777.1135807},
 doi = {10.1145/1135777.1135807},
 acmid = {1135807},
 publisher = {ACM},
 location = {New York, NY, USA}
}

@inproceedings{Nallapati:2008,
 author = {Nallapati, Ramesh M. and Ahmed, Amr and Xing, Eric P. and Cohen, William W.},
 title = {Joint Latent Topic Models for Text and Citations},
 booktitle = {Proceedings of the 14th ACM SIGKDD International Conference on Knowledge Discovery and Data Mining},
 series = {KDD '08},
 year = {2008},
 isbn = {978-1-60558-193-4},
 location = {Las Vegas, Nevada, USA},
 pages = {542--550},
 numpages = {9},
 url = {http://doi.acm.org/10.1145/1401890.1401957},
 doi = {10.1145/1401890.1401957},
 acmid = {1401957},
 publisher = {ACM}
}

@inproceedings{Cohn:2000,
 author = {Cohn, David and Hofmann, Thomas},
 title = {The Missing Link: A Probabilistic Model of Document Content and Hypertext Connectivity},
 booktitle = {Proceedings of the 13th International Conference on Neural Information Processing Systems},
 series = {NIPS'00},
 year = {2000},
 location = {Denver, CO},
 pages = {409--415},
 numpages = {7},
 url = {http://dl.acm.org/citation.cfm?id=3008751.3008811},
 acmid = {3008811},
 publisher = {MIT Press},
}

@inproceedings{Yu2017WSDM,
 author = {Yu, Wenchao and Aggarwal, Charu C. and Wang, Wei},
 title = {Temporally Factorized Network Modeling for Evolutionary Network Analysis},
 booktitle = {Proceedings of the Tenth ACM International Conference on Web Search and Data Mining},
 series = {WSDM '17},
 year = {2017},
 isbn = {978-1-4503-4675-7},
 location = {Cambridge, United Kingdom},
 pages = {455--464},
 numpages = {10},
 url = {http://doi.acm.org/10.1145/3018661.3018669},
 doi = {10.1145/3018661.3018669},
 acmid = {3018669},
 publisher = {ACM}
}

@inproceedings{lin2008facetnet,
  title={Facetnet: a framework for analyzing communities and their evolutions in dynamic networks},
  author={Lin, Yu-Ru and Chi, Yun and Zhu, Shenghuo and Sundaram, Hari and Tseng, Belle L},
  booktitle={Proceedings of the 17th international conference on World Wide Web},
  pages={685--694},
  year={2008},
  organization={ACM}
}

@inproceedings{chakrabarti2006evolutionary,
  title={Evolutionary clustering},
  author={Chakrabarti, Deepayan and Kumar, Ravi and Tomkins, Andrew},
  booktitle={Proceedings of the 12th ACM SIGKDD international conference on Knowledge discovery and data mining},
  pages={554--560},
  year={2006},
  organization={ACM}
}

@article{kim2009particle,
  title={A particle-and-density based evolutionary clustering method for dynamic networks},
  author={Kim, Min-Soo and Han, Jiawei},
  journal={Proceedings of the VLDB Endowment},
  volume={2},
  number={1},
  pages={622--633},
  year={2009},
  publisher={VLDB Endowment}
}

@article{kawadia2012sequential,
  title={Sequential detection of temporal communities by estrangement confinement},
  author={Kawadia, Vikas and Sreenivasan, Sameet},
  journal={Scientific reports},
  volume={2},
  year={2012},
  publisher={Nature Publishing Group}
}

@article{cohn2001missing,
  title={The missing link-a probabilistic model of document content and hypertext connectivity},
  author={Cohn, David and Hofmann, Thomas},
  journal={Advances in neural information processing systems},
  pages={430--436},
  year={2001},
  publisher={MIT; 1998}
}

@inproceedings{nallapati2008joint,
  title={Joint latent topic models for text and citations},
  author={Nallapati, Ramesh M and Ahmed, Amr and Xing, Eric P and Cohen, William W},
  booktitle={Proceedings of the 14th ACM SIGKDD international conference on Knowledge discovery and data mining},
  pages={542--550},
  year={2008},
  organization={ACM}
}

@article{zhou2009graph,
  title={Graph clustering based on structural/attribute similarities},
  author={Zhou, Yang and Cheng, Hong and Yu, Jeffrey Xu},
  journal={Proceedings of the VLDB Endowment},
  volume={2},
  number={1},
  pages={718--729},
  year={2009},
  publisher={VLDB Endowment}
}

@article{watts2002identity,
  title={Identity and search in social networks},
  author={Watts, Duncan J and Dodds, Peter Sheridan and Newman, Mark EJ},
  journal={science},
  volume={296},
  number={5571},
  pages={1302--1305},
  year={2002},
  publisher={American Association for the Advancement of Science}
}

@article{ravasz2002hierarchical,
  title={Hierarchical organization of modularity in metabolic networks},
  author={Ravasz, Erzs{\'e}bet and Somera, Anna Lisa and Mongru, Dale A and Oltvai, Zolt{\'a}n N and Barab{\'a}si, A-L},
  journal={science},
  volume={297},
  number={5586},
  pages={1551--1555},
  year={2002},
  publisher={American Association for the Advancement of Science}
}

@inproceedings{Ruan:2013,
 author = {Ruan, Yiye and Fuhry, David and Parthasarathy, Srinivasan},
 title = {Efficient Community Detection in Large Networks Using Content and Links},
 booktitle = {Proceedings of the 22Nd International Conference on World Wide Web},
 series = {WWW '13},
 year = {2013},
 isbn = {978-1-4503-2035-1},
 location = {Rio de Janeiro, Brazil},
 pages = {1089--1098},
 numpages = {10},
 url = {http://doi.acm.org/10.1145/2488388.2488483},
 doi = {10.1145/2488388.2488483},
 acmid = {2488483},
 publisher = {ACM}
}

@inproceedings{Yang:2009,
 author = {Yang, Tianbao and Jin, Rong and Chi, Yun and Zhu, Shenghuo},
 title = {Combining Link and Content for Community Detection: A Discriminative Approach},
 booktitle = {Proceedings of the 15th ACM SIGKDD International Conference on Knowledge Discovery and Data Mining},
 series = {KDD '09},
 year = {2009},
 isbn = {978-1-60558-495-9},
 location = {Paris, France},
 pages = {927--936},
 numpages = {10},
 url = {http://doi.acm.org/10.1145/1557019.1557120},
 doi = {10.1145/1557019.1557120},
 acmid = {1557120},
 publisher = {ACM}
}

@inproceedings{Pietilanen:2012,
 author = {Pietil\"{a}nen, Anna-Kaisa and Diot, Christophe},
 title = {Dissemination in Opportunistic Social Networks: The Role of Temporal Communities},
 booktitle = {Proceedings of the Thirteenth ACM International Symposium on Mobile Ad Hoc Networking and Computing},
 series = {MobiHoc '12},
 year = {2012},
 isbn = {978-1-4503-1281-3},
 location = {Hilton Head, South Carolina, USA},
 pages = {165--174},
 numpages = {10},
 url = {http://doi.acm.org/10.1145/2248371.2248396},
 doi = {10.1145/2248371.2248396},
 acmid = {2248396},
 publisher = {ACM}
}

@inproceedings{liu2015community,
  title={Community Detection Based on Structure and Content: A Content Propagation Perspective},
  author={Liu, Liyuan and Xu, Linli and Wangy, Zhen and Chen, Enhong},
  booktitle={Data Mining (ICDM), 2015 IEEE International Conference on},
  pages={271--280},
  year={2015},
  organization={IEEE}
}

@inproceedings{tang2011dynamic,
  title={Dynamic community detection with temporal dirichlet process},
  author={Tang, Xuning and Yang, Christopher C},
  booktitle={Privacy, Security, Risk and Trust (PASSAT) and 2011 IEEE Third Inernational Conference on Social Computing (SocialCom), 2011 IEEE Third International Conference on},
  pages={603--608},
  year={2011},
  organization={IEEE}
}

@article{chen2013detecting,
  title={Detecting overlapping temporal community structure in time-evolving networks},
  author={Chen, Yudong and Kawadia, Vikas and Urgaonkar, Rahul},
  journal={arXiv preprint arXiv:1303.7226},
  year={2013}
}

@inproceedings{xu2014exploiting,
  title={Exploiting Paper Contents and Citation Links to Identify and Characterise Specialisations},
  author={Xu, Han and Martin, Eric and Mahidadia, Ashesh},
  booktitle={2014 IEEE International Conference on Data Mining Workshop},
  pages={613--620},
  year={2014},
  organization={IEEE}
}

@article{bazzi2016community,
  title={Community detection in temporal multilayer networks, with an application to correlation networks},
  author={Bazzi, Marya and Porter, Mason A and Williams, Stacy and McDonald, Mark and Fenn, Daniel J and Howison, Sam D},
  journal={Multiscale Modeling \& Simulation},
  volume={14},
  number={1},
  pages={1--41},
  year={2016},
  publisher={SIAM}
}

@inproceedings{Leskovec2008,
 author = {Leskovec, Jure and Lang, Kevin J. and Dasgupta, Anirban and Mahoney, Michael W.},
 title = {Statistical Properties of Community Structure in Large Social and Information Networks},
 booktitle = {Proceedings of the 17th International Conference on World Wide Web},
 series = {WWW '08},
 year = {2008},
 isbn = {978-1-60558-085-2},
 location = {Beijing, China},
 pages = {695--704},
 numpages = {10},
 url = {http://doi.acm.org/10.1145/1367497.1367591},
 doi = {10.1145/1367497.1367591},
 acmid = {1367591},
 publisher = {ACM}
}

@article{fortunato2010community,
  title={Community detection in graphs},
  author={Fortunato, Santo},
  journal={Physics reports},
  volume={486},
  number={3},
  pages={75--174},
  year={2010},
  publisher={Elsevier}
}

@article{girvan2002community,
  title={Community structure in social and biological networks},
  author={Girvan, Michelle and Newman, Mark EJ},
  journal={Proceedings of the national academy of sciences},
  volume={99},
  number={12},
  pages={7821--7826},
  year={2002},
  publisher={National Acad Sciences}
}

@article{Barnes1982,
author = {Earl R. Barnes},
title = {An Algorithm for Partitioning the Nodes of a Graph},
journal = {SIAM Journal on Algebraic Discrete Methods},
volume = {3},
number = {4},
pages = {541-550},
year = {1982},
doi = {10.1137/0603056},
URL = { http://dx.doi.org/10.1137/0603056},
eprint = { http://dx.doi.org/10.1137/0603056}
}

@inproceedings{bergstra2013making,
  title={Making a science of model search: Hyperparameter optimization in hundreds of dimensions for vision architectures},
  author={Bergstra, James and Yamins, Daniel and Cox, David},
  booktitle={International Conference on Machine Learning},
  pages={115--123},
  year={2013}
}

@article{shahriari2016taking,
  title={Taking the human out of the loop: A review of bayesian optimization},
  author={Shahriari, Bobak and Swersky, Kevin and Wang, Ziyu and Adams, Ryan P and De Freitas, Nando},
  journal={Proceedings of the IEEE},
  volume={104},
  number={1},
  pages={148--175},
  year={2016},
  publisher={IEEE}
}

@article{rousseeuw1987silhouettes,
  title={Silhouettes: a graphical aid to the interpretation and validation of cluster analysis},
  author={Rousseeuw, Peter J},
  journal={Journal of computational and applied mathematics},
  volume={20},
  pages={53--65},
  year={1987},
  publisher={Elsevier}
}

\end{document}